\documentclass{article}
\usepackage{ijcai18}

\usepackage{times}

\usepackage{stmaryrd}
\usepackage{amsfonts}
\usepackage{amssymb}
\usepackage{graphicx}
\usepackage{array,multirow}
\usepackage{boxedminipage}
\usepackage{mathtools}
\usepackage{fixltx2e}
\usepackage{textcomp}
\usepackage{url}
\usepackage{algorithm}
\usepackage{algorithmicx}
\usepackage{algpseudocode}
\usepackage{subcaption}

\usepackage[font={small}, margin=0.3cm]{caption}
\renewcommand\vec{\boldsymbol}
\algdef{SE}[SUBALG]{Indent}{EndIndent}{}{\algorithmicend\ }%
\algtext*{Indent}
\algtext*{EndIndent}
\newcommand{\rpm}{\sbox0{$1$}\sbox2{$\scriptstyle\pm$}
  \raise\dimexpr(\ht0-\ht2)/2\relax\box2 }
\title{An Adaptive Clipping Approach for Proximal Policy Optimization}
\author{Gang Chen\\
School of Engineering \\ and Computer Science, \\ Victoria University of Wellington,\\ Wellington, New Zealand  \\
aaron.chen@ecs.vuw.ac.nz
\And
Yiming Peng\\
School of Engineering \\ and Computer Science,\\  Victoria University of Wellington, \\ Wellington, New Zealand  \\
yiming.peng@ecs.vuw.ac.nz
\And
Mengjie Zhang\\
School of Engineering \\ and Computer Science, \\ Victoria University of Wellington, \\ Wellington, New Zealand  \\
mengjie.zhang@ecs.vuw.ac.nz}

\begin{document}

\maketitle

\begin{abstract}
Very recently proximal policy optimization (PPO) algorithms have been proposed as first-order optimization methods for effective reinforcement learning. While PPO is inspired by the same learning theory that justifies trust region policy optimization (TRPO), PPO substantially simplifies algorithm design  and improves data efficiency by performing multiple epochs of \emph{clipped policy optimization} from sampled data. Although clipping in PPO stands for an important new mechanism for efficient and reliable policy update, it may fail to adaptively improve learning performance in accordance with the importance of each sampled state. To address this issue, a new surrogate learning objective featuring an adaptive clipping mechanism is proposed in this paper, enabling us to develop a new algorithm, known as PPO-$\lambda$. PPO-$\lambda$ optimizes policies repeatedly based on a theoretical target for adaptive policy improvement. Meanwhile, destructively large policy update can be effectively prevented through both clipping and adaptive control of a hyperparameter $\lambda$ in PPO-$\lambda$, ensuring high learning reliability. PPO-$\lambda$ enjoys the same simple and efficient design as PPO. Empirically on several Atari game playing tasks and benchmark control tasks, PPO-$\lambda$ also achieved clearly better performance than PPO.
\end{abstract}

\section{Introduction}
\label{sec-intro}

Driven by the explosive interest from the research community and industry, reinforcement learning (RL) technologies are being advanced at an accelerating pace \cite{barreto2017,nachum2017,tang2017,haarnoja2017,sasha2017feudal}. Particularly, numerous innovative RL algorithms have been proposed in recent years for training deep neural networks (DNN) that can solve difficult decision making problems modelled as Markov Decision Processes (MDPs) \cite{arulkumaran2017,wang2015,Duan2016,minh2015}. Among them, many cutting-edge RL algorithms feature the use of policy search (PS) techniques where action-selection policies are represented explicitly as DNNs rather than implicitly derived from separate value functions. Through direct optimization of policy parameters in DNNs, these algorithms are expected to achieve near-optimal learning performance in extremely large learning environments \cite{minh2016,hausknecht2016,gu2017,schulman2015a,schulman2015,lillicrap2015,wang2016,wu2017}.

Trust region policy optimization (TRPO) is a prominent and highly effective algorithm for PS  \cite{schulman2015}. It has the goal of optimizing a surrogate learning objective subject to predefined behavioral constraints (see Subsection \ref{sub-trpo} for more information). To tackle this learning problem, both linear approximation of the learning objective and quadratic approximation of the constraint are utilized to jointly guide policy update, resulting in relatively high computation complexity. Meanwhile, TRPO is not compatible with NN architectures for parameter sharing (between the policy and value function) \cite{schulman2017}. It is also not suitable for performing multiple epochs of minibatch policy update which is vital for high sample efficiency.

In order to address the above difficulties, very recently Schulman \emph{et al.} developed new PS algorithms called proximal policy optimization (PPO) \cite{schulman2017}. By employing only first-order optimization methods, PPO is much simpler to implement and more efficient to operate than TRPO. Moreover, PPO allows repeated policy optimization based on previously sampled data to reduce sample complexity. Thanks to this clever design, on many benchmark RL tasks including robotic locomotion and Atari game playing, PPO significantly outperformed TRPO and several other state-of-the-art learning algorithms including A2C \cite{minh2016}, Vanilla PG \cite{sutton2000,bhatnagar2009} and CEM \cite{szita2006}. It also performed competitively to ACER, a currently leading algorithm in RL research \cite{wang2016}.

A simple clipping mechanism plays a vital role for reliable PS in PPO. In fact, clipping, as an efficient and effective technique for policy learning under behavioral restriction, is general in nature and can be widely used in many cutting-edge RL algorithms. However the clipped probability ratio used by PPO in its surrogate learning objective may allow less important states to receive more policy updates than desirable. This is because policy update at more important states often vanish early during repeated policy optimization whenever the corresponding probability ratios shoot beyond a given clipping threshold. Moreover, the updating scale at any sampled state remains fixed even when the policy behavior has been changed significantly at that state, potentially affecting learning performance too.

In view of the above issues, this paper aims to propose a new adaptive approach for clipped policy optimization. For this purpose, using the learning theory originally introduced by TRPO in \cite{schulman2015}, we will consider a learning problem at the level of each individual state. After transforming the problem into a Lagrangian and identifying its stationary point, the target for adaptive policy improvement can be further obtained. Guided by this target, we will propose a new surrogate learning objective featuring an adaptive clipping mechanism controlled by a hyperparameter $\lambda$, enabling us to develop a new RL algorithm called PPO-$\lambda$.

PPO-$\lambda$ is equally simple and efficient as PPO. We will investigate the performance of PPO-$\lambda$ through experiments. Since PPO-$\lambda$ finds its root in PPO which has already been demonstrated to outperform many state-of-the-art algorithms \cite{schulman2017}, we will focus on comparing PPO-$\lambda$ with PPO in this paper. Particularly, as evidenced on several Atari game playing tasks and benchmark control tasks in Section \ref{sec-empirical_analysis}, PPO-$\lambda$ can achieve clearly better performance and sample efficiency than PPO. Our results suggest that, in order to solve many difficult RL problems, PPO-$\lambda$ may be considered as a useful alternative to PPO.

\section{Background}
\label{sec-back}

Our discussion of the research background begins with a definition of the RL problem. We then briefly introduce TRPO and its supporting learning theory. Afterwards, the surrogate learning objective for PPO will be presented and analyzed.

\subsection{Reinforcement Learning Problem}
\label{sub-rl-prob}

The environment for RL is often described as a MDP with a set of states $\vec{s}\in \mathbb{S}$ and a set of actions $a\in \mathbb{A}$ \cite{sutton1998}. At each time step $t$, an RL agent observes its current state $\vec{s}_t$ and performs a selected actions $a_t$, resulting in a transition to a new state $\vec{s}_{t+1}$. Meanwhile, a scalar reward $r(\vec{s}_t,a_t)$ will be produced as an environmental feedback. Starting from arbitrary initial state $\vec{s}_0$ at time $t=0$, the goal for RL is to maximize the expected long-term cumulative reward defined below
\begin{equation}
V(\vec{s})=E\left\{ \sum_{t=0}^{\infty} \gamma^t r(\vec{s}_t,a_t) \left| \vec{s}_0=\vec{s} \right. \right\}
\label{equ-cum-reward}
\end{equation}

\noindent
where $\gamma$ is a discount factor. To fulfill this goal, an agent must identify a suitable policy $\pi$ for action selection. Specifically, $\pi_{\vec{\theta}}(\vec{s}_t,a_t)$ is often treated as a parametric function that dictates the probability of choosing any action $a_t$ at every possible state $\vec{s}_t$, controlled by parameters $\vec{\theta}$. Subsequently, $V(\vec{s})$ in (\ref{equ-cum-reward}) becomes a function of $\vec{\theta}$ since an agent's behavior depends completely on its policy $\pi_{\vec{\theta}}$. Hence the problem for RL is to learn the optimal policy parameters, i.e. $\vec{\theta}^*$:
\begin{equation}
\vec{\theta}^*=\operatorname{arg\,max}_{\vec{\theta}} V^{\pi_{\vec{\theta}}}
\label{equ-opt-theta}
\end{equation}

\noindent
where $V^{\pi_{\vec{\theta}}}$ measures the expected performance of using policy $\pi_{\vec{\theta}}$ and is defined below
$$
V^{\pi_{\vec{\theta}}}=\sum_{\vec{s}_0} \rho_{0}(\vec{s}_0) V^{\pi_{\vec{\theta}}}(\vec{s}_0)
$$

\noindent
with $\rho_0(\vec{s}_0)$ representing the probability for the RL agent to start interacting with its environment at state $\vec{s}_0$. $V^{\pi_{\vec{\theta}}}(\vec{s}_0)$ stands for the expected cumulative reward obtainable upon following policy $\pi_{\vec{\theta}}$ from state $\vec{s}_0$.

\subsection{Trust Region Bound and Policy Optimization}
\label{sub-trpo}

Instead of maximizing $V^{\pi_{\vec{\theta}}}$ directly, TRPO is designed to optimize a performance lower bound. To find it, we need to define the state-action value function $Q^{\pi_{\vec{\theta}}}$ and the advantage function $A^{\pi_{\vec{\theta}}}$ with respect to any policy $\pi_{\vec{\theta}}$ below
\begin{equation}
Q^{\pi_{\vec{\theta}}}(\vec{s},a)=E\left\{ \sum_{t=0}^{\infty} \gamma^t r(\vec{s}_t,a_t) \left| \vec{s}_0=\vec{s}, a_0=a \right. \right\}
\label{equ-q-func}
\end{equation}
\begin{equation}
A^{\pi_{\vec{\theta}}}(\vec{s},a)=Q^{\pi_{\vec{\theta}}}(\vec{s},a)-V^{\pi_{\vec{\theta}}}(\vec{s})
\label{equ-adv-func}
\end{equation}

\noindent
where $a_t \sim \pi_{\vec{\theta}}(\vec{s}_t,a_t)$ for $t\geq 1$. Further define function $L_{\pi_{\vec{\theta}}}(\pi_{\vec{\theta'}})$ as
\begin{equation}
L_{\pi_{\vec{\theta}}}(\pi_{\vec{\theta'}})=V^{\pi_{\vec{\theta}}}+\sum_{\vec{s}} \rho_{\pi_{\vec{\theta}}}(\vec{s}) \sum_{a} \pi_{\vec{\theta}'}(\vec{s},a) A^{\pi_{\vec{\theta}}}(\vec{s},a)
\label{equ-l-func}
\end{equation}

\noindent
Here $\rho_{\pi_{\vec{\theta}}}(\vec{s})$ refers to the discounted frequencies of visiting every state $\vec{s}$ while obeying policy $\pi_{\vec{\theta}}$ \cite{sutton2000}. Given any two policies $\pi_{\vec{\theta}_{old}}$ and $\pi_{\vec{\theta}_{new}}$. Assume that $\pi_{\vec{\theta}_{old}}$ is the policy utilized to sample state transition data and $\pi_{\vec{\theta}_{new}}$ stands for the policy obtained through optimization based on sampled data. Then the following performance bound holds \cite{schulman2015}:
\begin{equation}
V^{ \pi_{\vec{\theta}_{new}} } \geq L_{ \pi_{\vec{\theta}_{old}} } ( \pi_{\vec{\theta}_{new}} ) - \frac{ 4\epsilon \gamma }{ 1-\gamma^2 } \alpha^2
\label{equ-perf-bound}
\end{equation}

\noindent
with $\epsilon$ being a non-negative constant and
\begin{equation*}
\begin{split}
\alpha &= D_{TV}^* \\
& = \max_{\vec{s}} D_{TV}(\pi_{\vec{\theta}_{old}}(\vec{s},\cdot), \pi_{\vec{\theta}_{new}}(\vec{s},\cdot)) \\
& = \max_{\vec{s}} \frac{1}{2} \sum_{a} \left| \pi_{\vec{\theta}_{old}}(\vec{s},a) - \pi_{\vec{\theta}_{new}}(\vec{s},\cdot) \right|
\end{split}
\end{equation*}

\noindent
$D_{TV}$ is the total variation divergence. Provided that $\pi_{\vec{\theta}_{old}}$ and $\pi_{\vec{\theta}_{new}}$ are reasonably close, steady policy improvement can be guaranteed during RL based on the performance bound in (\ref{equ-perf-bound}). Since $D_{TV}^2(\pi_{\vec{\theta}_{old}}, \pi_{\vec{\theta}_{new}})\leq D_{KL} (\pi_{\vec{\theta}_{old}}, \pi_{\vec{\theta}_{new}})$ where $D_{KL}$ stands for the KL divergence \cite{pollard2000}, during each learning iteration, TRPO is designed to solve the following constrained optimization problem:
\begin{equation}
\max_{\vec{\theta}_{new}} L_{ \pi_{\vec{\theta}_{old}} } ( \pi_{\vec{\theta}_{new}} )\ s.t.\ \bar{D}_{KL}^{\pi_{\vec{\theta}_{old}}}(\pi_{\vec{\theta}_{old}}, \pi_{\vec{\theta}_{new}})\leq \delta
\label{equ-trpo-prob}
\end{equation}

\noindent
where $\bar{D}_{KL}^{\pi_{\vec{\theta}_{old}}}$ is the average KL divergence over all states to be visited upon following policy $\pi_{\vec{\theta}_{old}}$. It heuristically replaces the maximum $D_{KL}^*$ such that an approximated solution of (\ref{equ-trpo-prob}) can be obtained efficiently in TRPO. However it is to be noted that $D_{KL}^*$ should actually serve as the constraint in theory so as to guarantee continued policy improvement according to (\ref{equ-perf-bound}).

\subsection{Proximal Policy Optimization through a Clipped Surrogate Learning Objective}
\label{sub-ppo}

Because TRPO uses $\bar{D}_{KL}$ in (\ref{equ-trpo-prob}) to guide its policy update, the updated policy may exhibit strong behavioral deviations at some sampled states. In practice, it can be more desirable to clip policy optimization in response to large changes to the probability ratio $\tau_t(\pi_{\vec{\theta}_{new}})=\frac{ \pi_{\vec{\theta}_{new}}(\vec{s}_t,a_t) }{ \pi_{\vec{\theta}_{old}}(\vec{s}_t,a_t) }$. It is straightforward to see that $\tau_t$ applies to each state individually. For any state $s_t$, guarded by $\delta$ which is a given threshold on probability ratios, policy update at $s_t$ can be penalized through the clipping function below.
\begin{equation}
C(\vec{s}_t)=
\left\{
\begin{array}{cc}
  (1+\delta) A_t^{\pi_{\vec{\theta}_{old}} } & A_t^{\pi_{\vec{\theta}_{old}} } > 0,\ \tau_t>1+\delta \\
  (1-\delta) A_t^{\pi_{\vec{\theta}_{old}} } & A_t^{\pi_{\vec{\theta}_{old}} } < 0,\ \tau_t<1-\delta \\
  \tau_t(\pi_{\vec{\theta}_{new}}) A_t^{\pi_{\vec{\theta}_{old}} } & otherwise
\end{array}
\right.
\label{equ-ppo-clip}
\end{equation}

\noindent
Here for simplicity we use $A_t^{\pi_{\vec{\theta}_{old}} }$ to denote $A^{\pi_{\vec{\theta}_{old}}}(\vec{s}_t,a_t)$. Upon optimizing $C(\vec{s}_t)$ in (\ref{equ-ppo-clip}) as the surrogate learning objective, penalties apply as long as either of the first two conditions in (\ref{equ-ppo-clip}) are satisfied, in which case $|\tau_t-1|>\delta$ and policy behavioral changes at $\vec{s}_t$ are hence considered to be too large. Building on $C(\vec{s}_t)$, PPO can effectively prevent destructively large policy updates and has been shown to achieve impressive learning performance empirically \cite{schulman2017}.

It is not difficult to see that $A_t^{\pi_{\vec{\theta}_{old}} }$ often varies significantly at different sampled states $\vec{s}_t$. In particular states with comparatively large values of $|A_t^{\pi_{\vec{\theta}_{old}} }|$ are deemed as important since policy updates at those states could potentially lead to high performance improvements. However, through repeated updating of policy parameters along the direction of $\partial C(\vec{s}_t)/\partial \vec{\theta}_{new}$, $\tau_t(\pi_{\vec{\theta}_{new}})$ at some important states $\vec{s}_t$ can easily go beyond the region $(1-\delta,1+\delta)$ and hence either of the first two cases in (\ref{equ-ppo-clip}) may be satisfied. In such situation, $\partial C(\vec{s}_t)/\partial \vec{\theta}_{new}=0$ and policy update at state $\vec{s}_t$ vanishes. In comparison, policy updates at less important states are less likely to vanish.

Due to the above reason, after multiple epochs of policy update in a learning iteration (see the PPO algorithm in \cite{schulman2017}), less important states tend to have been updated more often than important states. As a consequence, the scale of policy updates may not properly reflect the importance of each sampled state. Meanwhile, we note that, whenever policy update does not vanish, $\partial C(\vec{s}_t)/\partial \vec{\theta}_{new}$ is determined by $A_t^{\pi_{\vec{\theta}_{old}} }$ which remains fixed during a full learning iteration. In other words, the scale of policy update stays at the same level across multiple updating epochs. This may not be desirable either since, when $\pi_{\vec{\theta}_{new}}$ deviates significantly from $\pi_{\vec{\theta}_{old}}$, further policy updates often result in undesirable outcome. Ideally, with every new epoch of successive policy update, we expect the updating scale to be adaptively reduced.

Motivated by our analysis of $C(\vec{s}_t)$ and PPO, this paper aims to show that, by employing a new adaptive clipping method for constrained policy optimization, we can effectively improve PS performance and sample efficiency.

\section{A New Adaptive Approach for Clipped Policy Optimization}
\label{sec-ppo-lambda}

While developing the new adaptive approach for clipped policy updating, we wish to achieve three objectives: (O1) similar to PPO, policy behavioral changes should be controlled at the level of individual states; (O2) in comparison to important states, the policy being learned is expected to exhibit less changes at less important states; and (O3) the policy updating scale should be reduced adaptively with each successive epoch in a learning iteration. Driven by the three objectives, we decide to study a learning problem slightly different from that considered by TRPO. For this purpose, we note that, although $D_{KL}(\pi_{\vec{\theta}_{old}},\pi_{\vec{\theta}_{new}})\neq D_{KL}(\pi_{\vec{\theta}_{new}},\pi_{\vec{\theta}_{old}})$, it is straightforward to see that $D_{KL}(\pi_{\vec{\theta}_{new}},\pi_{\vec{\theta}_{old}}) > D^2_{TV}(\pi_{\vec{\theta}_{new}},\pi_{\vec{\theta}_{old}})=D^2_{TV}(\pi_{\vec{\theta}_{old}},\pi_{\vec{\theta}_{new}})$. Therefore, the performance bound in (\ref{equ-perf-bound}) remains valid with
$$
\alpha=D^*_{KL}=\max_{\vec{s}} D_{KL}\left( \pi_{\vec{\theta}_{new}}(\vec{s},\cdot),\pi_{\vec{\theta}_{old}}(\vec{s},\cdot) \right)
$$

Hence if the maximum KL divergence across all states can be bounded properly, we can guarantee policy improvement in each learning iteration. In view of this and objective O1, it is sensible to concentrate on local policy optimization at every sampled state first and then combine policy updates from all sampled states together. At any specific state $\vec{s}$, maximizing $L_{\pi_{\vec{\theta}_{old}}}(\pi_{\vec{\theta}_{new}})$ in (\ref{equ-l-func}) is equivalent to maximizing $\sum_a \pi_{\vec{\theta}_{new}}(\vec{s},a) A^{\pi_{\vec{\theta}_{old}}}$. As a consequence, the learning problem for PPO-$\lambda$ at state $\vec{s}$ can be defined as
\begin{equation}
\begin{split}
& \max_{\vec{\theta}_{new}} \sum_a \pi_{\vec{\theta}_{new}}(\vec{s},a) A^{\pi_{\vec{\theta}_{old}}} (\vec{s},a)\\
& s.t.\ D_{KL}(\pi_{\vec{\theta}_{new}}(\vec{s},\cdot), \pi_{\vec{\theta}_{old}}(\vec{s},\cdot))\leq \delta
\end{split}
\label{equ-ppo-prob}
\end{equation}

\noindent
Different from TRPO that considers $D_{KL}(\pi_{\vec{\theta}_{old}}, \pi_{\vec{\theta}_{new}})$ as its optimization constraint, we consider $D_{KL}(\pi_{\vec{\theta}_{new}}, \pi_{\vec{\theta}_{old}})$ in (\ref{equ-ppo-prob}) instead. The Lagrangian of our learning problem is
\begin{equation}
\begin{split}
\mathcal{L}= & \sum_a \pi_{\vec{\theta}_{new}}(\vec{s},a) A^{\pi_{\vec{\theta}_{old}}}(\vec{s},a)\\
& -\lambda \left( \sum_a \pi_{\vec{\theta}_{new}}(\vec{s},a) \log \frac{\pi_{\vec{\theta}_{new}}(\vec{s},a)}{\pi_{\vec{\theta}_{old}}(\vec{s},a)} - \delta \right)
\end{split}
\label{equ-lagrange}
\end{equation}

\noindent
where the hyperparameter $\lambda$ stands for the Lagrange multiplier. It controls the penalty due to the violation of the local policy optimization constraint. According to \cite{gelfand2000}, we can identify the stationary point of $\mathcal{L}$ by solving the following Euler-Lagrangian equation:
\begin{equation}
\begin{split}
\frac{\partial \mathcal{L}}{\partial \pi_{\vec{\theta}_{new}}(\vec{s},a)} &= A^{\pi_{\vec{\theta}_{old}}} (\vec{s},a) - \lambda - \lambda \log \pi_{\vec{\theta}_{new}}(\vec{s},a) \\
& + \lambda \log \pi_{\vec{\theta}_{old}}(\vec{s},a) \\
&=0
\end{split}
\label{equ-el-equa}
\end{equation}

\noindent
Under the conditions that $\forall \vec{s}\in\mathbb{S}$:
$$
\sum_a \pi_{\vec{\theta}_{old}}(\vec{s},a)=1,\ \sum_a \pi_{\vec{\theta}_{new}}(\vec{s},a)=1
$$

The equation in (\ref{equ-el-equa}) can be solved immediately. As a matter of fact, it can be easily verified that
\begin{equation}
\pi^*_{\vec{\theta}_{new}}(\vec{s},a)\propto \pi_{\vec{\theta}_{old}}(\vec{s},a) e^{ \frac{A^{\pi_{\vec{\theta}_{old}}}(\vec{s},a)}{\lambda} }
\label{equ-el-sol}
\end{equation}

\noindent
where $\pi^*_{\vec{\theta}_{new}}$ represents the target for adaptive policy improvement in PPO-$\lambda$. To guide the updating of $\pi_{\vec{\theta}_{new}}$ towards the corresponding learning target $\pi^*_{\vec{\theta}^*_{new}}$ in (\ref{equ-el-sol}), we can quantify the difference in between $\pi_{\vec{\theta}_{new}}$ and $\pi^*_{\vec{\theta}^*_{new}}$ through KL divergence as
\begin{equation}
\begin{split}
& D_{KL}(\pi_{\vec{\theta}_{new}} (\vec{s},\cdot), \pi^*_{\vec{\theta}_{new}} (\vec{s},\cdot))= \\
& \sum_a \pi_{\vec{\theta}_{new}} (\vec{s},a) \log \frac{ \pi_{\vec{\theta}_{new}} (\vec{s},a) }{ \pi^*_{\vec{\theta}_{new}} (\vec{s},a) }
\end{split}
\label{equ-kl-target}
\end{equation}

For the purpose of minimizing $D_{KL}(\pi_{\vec{\theta}_{new}}, \pi^*_{\vec{\theta}_{new}})$ in (\ref{equ-kl-target}) through updating $\vec{\theta}_{new}$, the derivative below can be utilized by the PPO-$\lambda$ algorithm.
\begin{equation}
\frac{\partial D_{KL}(\pi_{\vec{\theta}_{new}}, \pi^*_{\vec{\theta}_{new}}) }{\partial \vec{\theta}_{new}} = \sum_a \frac{\partial \pi_{\vec{\theta}_{new}} (\vec{s},a) }{\partial \vec{\theta}_{new}} \log \frac{ \pi_{\vec{\theta}_{new}} (\vec{s},a) }{ \pi^*_{\vec{\theta}_{new}} (\vec{s},a) }
\label{equ-kl-derv}
\end{equation}

\noindent
Because the summation over all actions $a\in\mathbb{A}$ in (\ref{equ-kl-derv}) can be cumbersome to perform in practice, especially when the RL problem allows a huge collection of alternative actions, we choose to adopt the same importance sampling technique used by TRPO such that
\begin{equation}
\begin{split}
\frac{\partial D^t_{KL} }{\partial \vec{\theta}_{new} } & \approx \frac{1}{ \pi_{\vec{\theta}_{old}}(\vec{s}_t,a_t) } \frac{\partial \pi_{\vec{\theta}_{new}} (\vec{s}_t,a_t) }{\partial \vec{\theta}_{new}} \log \frac{ \pi_{\vec{\theta}_{new}} (\vec{s}_t,a_t) }{ \pi^*_{\vec{\theta}_{new}} (\vec{s}_t,a_t) } \\
& = \frac{\partial \tau_t }{\partial \vec{\theta}_{new}  } \log \frac{ \pi_{\vec{\theta}_{new}} (\vec{s}_t,a_t) }{ \pi^*_{\vec{\theta}_{new}} (\vec{s}_t,a_t) }
\end{split}
\label{equ-kl-approx-derv}
\end{equation}

\noindent
where $D^t_{KL}$ refers to the KL divergence between $\pi_{\vec{\theta}_{new}}$ and $\pi^*_{\vec{\theta}_{new}}$ at time $t$ with sampled state $\vec{s}_t$ and action $a_t$. It is interesting to note that the same technique is also utilized in PPO, which updates $\vec{\theta}_{new}$ at any time $t$ (ignoring value function and entropy related update for the moment) in the direction of
\begin{equation}
\frac{\partial \tau_t }{\partial \vec{\theta}_{new}  } A^{\pi_{\vec{\theta}_{old}}}(\vec{s}_t,a_t)
\label{equ-ppo-updt}
\end{equation}

\noindent
Notice that, in the first updating epoch of every learning iteration (see Algorithm \ref{alg-1}), because $\vec{\theta}_{new}=\vec{\theta}_{old}$, we can see that,
\begin{equation}
\frac{\partial D^t_{KL} }{\partial \vec{\theta}_{new} } \approx -\frac{1}{\lambda} \frac{\partial \tau_t }{\partial \vec{\theta}_{new}  } A^{\pi_{\vec{\theta}_{old}}}(\vec{s}_t,a_t)
\label{equ-kl-derv-eqv}
\end{equation}

\noindent
In view of (\ref{equ-ppo-updt}) and (\ref{equ-kl-derv-eqv}), to ensure that the updating step size of PPO-$\lambda$ is at the same level as PPO, it is natural for us to update $\vec{\theta}_{new}$ according to $\lambda \frac{\partial D^t_{KL} }{\partial \vec{\theta}_{new} }$ in PPO-$\lambda$. Meanwhile, we must penalize large policy updates through clipping. Therefore, we propose to adopt a new surrogate learning objective at any time $t$ as
\begin{equation}
\hat{D}_{KL}^t =
\lambda
\left\{
\begin{array}{cc}
  (1+\delta) \log \frac{ \pi^t_{\vec{\theta}_{new}} }{ \pi^{*,t}_{\vec{\theta}_{new}} } & A_t^{\pi_{\vec{\theta}_{old}} } > 0,\ \tau_t>1+\delta \\
  (1-\delta) \log \frac{ \pi^t_{\vec{\theta}_{new}} }{ \pi^{*,t}_{\vec{\theta}_{new}} } & A_t^{\pi_{\vec{\theta}_{old}} } < 0,\ \tau_t<1-\delta \\
  \tau_t(\pi_{\vec{\theta}_{new}}) \log \frac{ \pi^t_{\vec{\theta}_{new}} }{ \pi^{*,t}_{\vec{\theta}_{new}} } & otherwise
\end{array}
\right.
\label{equ-ppol-surr}
\end{equation}

In comparison to (\ref{equ-ppo-clip}), we can spot two advantages of using (\ref{equ-ppol-surr}). First, at a less important state $\vec{s}_t$, $\pi^{*,t}_{\vec{\theta}_{new}}/\pi^t_{\vec{\theta}_{old}}$ will fall within $(1-\delta,1+\delta)$ with large enough $\lambda$ in (\ref{equ-el-sol}). Consequently repeated updates at $\vec{s}_t$ will not push $\pi^t_{\vec{\theta}_{new}}$ towards the clipping boundary and are expected to produce relatively small policy behavioral changes. Due to this reason, objective O2 is achieved. Furthermore, after each updating epoch, we can expect $\pi^{t}_{\vec{\theta}_{new}}$ to become closer to $\pi^{*,t}_{\vec{\theta}_{new}}$, thereby reducing the absolute value of $\log (\pi^t_{\vec{\theta}_{new}}/ \pi^{*,t}_{\vec{\theta}_{new}})$ in (\ref{equ-ppol-surr}). Accordingly, policy update will adaptively reduce its scale with successive updating epochs. Objective O3 is hence realized.

While training a NN architecture that shares parameters between the policy and value function through PPO-$\lambda$, our surrogate learning objective must take into account a value function error term and an entropy bonus term \cite{schulman2017}. As a consequence, at any time $t$, $\vec{\theta}_{new}$ can be updated according to the learning rule below.
\begin{equation}
\begin{split}
\vec{\theta}_{new} \leftarrow & \vec{\theta}_{new} - \beta \lambda\frac{\partial \hat{D}^t_{KL} }{\partial \vec{\theta}_{new} } - \beta c_1 \frac{\partial E_V(\vec{s}_t)^2}{\partial \vec{\theta}_{new} } \\
& + \beta c_2 \frac{\partial S(\vec{s}_t) }{\partial \vec{\theta}_{new}}
\end{split}
\label{equ-ppol-lr}
\end{equation}

\noindent
where $c_1$ and $c_2$ are two fixed coefficients. $\beta$ is the learning rate. $E_V(\vec{s}_t)$ stands for the TD error of the value function at state $\vec{s}_t$. $S(\vec{s}_t)$ represents the entropy of the currently learned policy $\pi_{\vec{\theta}_{new}}$ at $\vec{s}_t$. The learning rule in (\ref{equ-ppol-lr}) requires estimation of the advantage function, i.e. $A^{\pi_{\vec{\theta}_{old}}}_t$. For this purpose, identical to PPO, PPO-$\lambda$ employs the generalized advantage function estimation technique introduced in \cite{schulman2015a}.

In practice, based on the learning rule in (\ref{equ-ppol-lr}), we will update policy parameters $\vec{\theta}_{new}$ with minibatch SGD over multiple time steps of data. This is directly achieved by using the Adam algorithm \cite{kingma2014}. In the mean time, we should prevent large behavioral deviation of $\pi_{\vec{\theta}_{new}}$ from $\pi_{\vec{\theta}_{old}}$ for reliable learning. Particularly, following the technique adopted by the PPO implementation in OpenAI baselines\footnote{https://www.github.com/openai/baselines}, the threshold on probability ratios $\delta$ is linearly decremented from the initial threshold of $\delta_0$ to $0$ when learning approaches to its completion. Accordingly, the learning rate $\beta$ also decrements linearly from the initial rate of $\beta_0$ to $0$. In line with this linear decrement strategy, $\lambda$ must be adjusted in every learning iteration to continually support objective O2. Notice that, in the PPO implementation, the estimated advantage function values will be normalized according to (\ref{equ-adv-norm}) at the beginning of each learning iteration \cite{schulman2017},
\begin{equation}
\tilde{A}^{\pi_{\vec{\theta}_{old}}}=\frac{ A^{\pi_{\vec{\theta}_{old}}} - \bar{A}^{\pi_{\vec{\theta}_{old}}}   }{ \sigma_{ A^{\pi_{\vec{\theta}_{old}}} } }
\label{equ-adv-norm}
\end{equation}

\noindent
where $\bar{A}^{\pi_{\vec{\theta}_{old}}}$ is the mean advantage function values in a minibatch and $\sigma_{ A^{\pi_{\vec{\theta}_{old}}} }$ gives the corresponding standard deviation. We assume that the normalized advantage function values in the minibatch follow the standard normal distribution. Subsequently, to ensure that the same proportion of sampled states will have their policy improvement target staying in the clipping boundary across all learning iterations, $\lambda$ will be determined as below
\begin{equation}
\lambda_n=\lambda_0 \frac{\log (\delta_0 + 1)}{\log (\delta_n+1)}
\label{equ-lam-adj}
\end{equation}

\noindent
with $\lambda_n$ and $\delta_n$ stand respectively for $\lambda$ and $\delta$ to be used in the $n$-th learning iteration. Our discussion above has been summarized in Algorithm \ref{alg-1}. It should be clear now that PPO-$\lambda$ achieves all the three objectives highlighted at the beginning of this section. In comparison to PPO, PPO-$\lambda$ is equally simple in design. It can be implemented easily by applying only minor changes to the PPO implementation. As for operation efficiency, PPO and PPO-$\lambda$ are indistinguishable too.

\begin{algorithm}[!ht]
 \begin{algorithmic}[1]
   \Require an MDP $\langle \mathbb{S}, \mathbb{A}, \mathbb{P}, \mathbb{R}, \gamma\rangle$,
   \Statex $n$: the number of learning iterations,
   \Statex $p$: the number of actors,
   \Statex $T$: the horizon,
   \Statex $K$: the number of epochs
   \Statex $M$: the minibatch size
   \Statex $\lambda$: the Lagrangian multiplier\\
    %\State Given $e$ episodes.
    \textbf{repeat} for $n$ learning iterations: \\
    \hspace{0.45cm}\textbf{repeat} for $p$ actors:
        \State \hspace{0.90cm} Perform a rollout with $\pi_{\vec{\theta}_{\text{old}}}$ for $T$ time steps
        \State \hspace{0.90cm} Compute $\hat{A}_{1}$,$\dots$,$\hat{A}_{T}$
        
        \State \hspace{0.40cm} Compute $\lambda$ according to \eqref{equ-lam-adj} \\
        \hspace{0.45cm}\textbf{repeat} for $k$ epochs:
            \State \hspace{0.90cm} Perform minibatch SGD update of $\vec{\theta}_{\text{new}}$ based on

                       \ \ \  (\ref{equ-ppol-lr}) and $M$ samples \\
                        \hspace{0.90cm} Repeat step 7 until $p\times T$ samples have been used
        \State \hspace{0.40cm} $\vec{\theta}_{\text{old}} \leftarrow \vec{\theta}_{\text{new}} $
        % \EndIntent
    %\Return $\vec{\theta}$
 \end{algorithmic}
\caption{PPO-$\lambda$}
\label{alg-1}
\end{algorithm}

\begin{figure}[!ht]
      \centering
      \begin{minipage}[t]{0.085\textwidth}
        \centering
        \includegraphics[width=\textwidth]{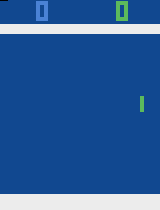}
        \subcaption{Step 1}
      \end{minipage}%
      \hspace{0.015cm}
      \begin{minipage}[t]{0.085\textwidth}
        \centering
        \includegraphics[width=\textwidth]{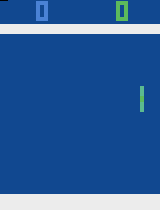}
        \subcaption{Step 2}
      \end{minipage}
      \begin{minipage}[t]{0.085\textwidth}
        \centering
        \includegraphics[width=\textwidth]{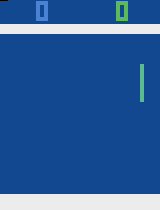}
        \subcaption{Step 3}
      \end{minipage}
      \begin{minipage}[t]{0.085\textwidth}
        \centering
        \includegraphics[width=\textwidth]{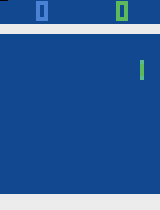}
        \subcaption{Step 4}
      \end{minipage}
      \begin{minipage}[t]{0.085\textwidth}
        \centering
        \includegraphics[width=\textwidth]{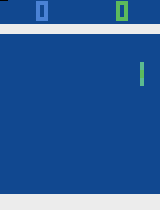}
        \subcaption{Step 5}
      \end{minipage}
      \caption{The snapshots of the first five steps taken in the Pong game when using the max pixel values of sampled frames.}
      \label{fig:max_values}
  \end{figure}
\begin{figure}[!ht]
      \centering
      \begin{minipage}[t]{0.085\textwidth}
        \centering
        \includegraphics[width=\textwidth]{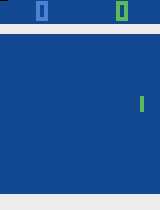}
        \subcaption{Step 1}
      \end{minipage}%
      \hspace{0.015cm}
      \begin{minipage}[t]{0.085\textwidth}
        \centering
        \includegraphics[width=\textwidth]{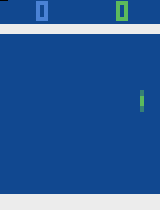}
        \subcaption{Step 2}
      \end{minipage}
      \begin{minipage}[t]{0.085\textwidth}
        \centering
        \includegraphics[width=\textwidth]{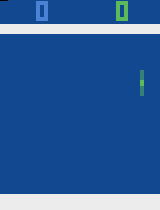}
        \subcaption{Step 3}
      \end{minipage}
      \begin{minipage}[t]{0.085\textwidth}
        \centering
        \includegraphics[width=\textwidth]{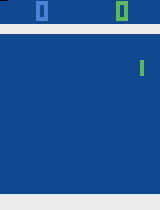}
        \subcaption{Step 4}
      \end{minipage}
      \begin{minipage}[t]{0.085\textwidth}
        \centering
        \includegraphics[width=\textwidth]{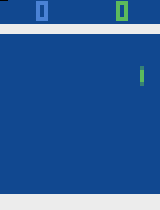}
        \subcaption{Step 5}
      \end{minipage}
      \caption{The snapshots of the first five steps taken in the Pong game when using the mean pixel values of sampled frames.}
      \label{fig:mean_values}
  \end{figure}
 \begin{figure*}[!t]
      \centering
      \begin{minipage}[t]{0.32\textwidth}
        \centering
        \includegraphics[width=\textwidth]{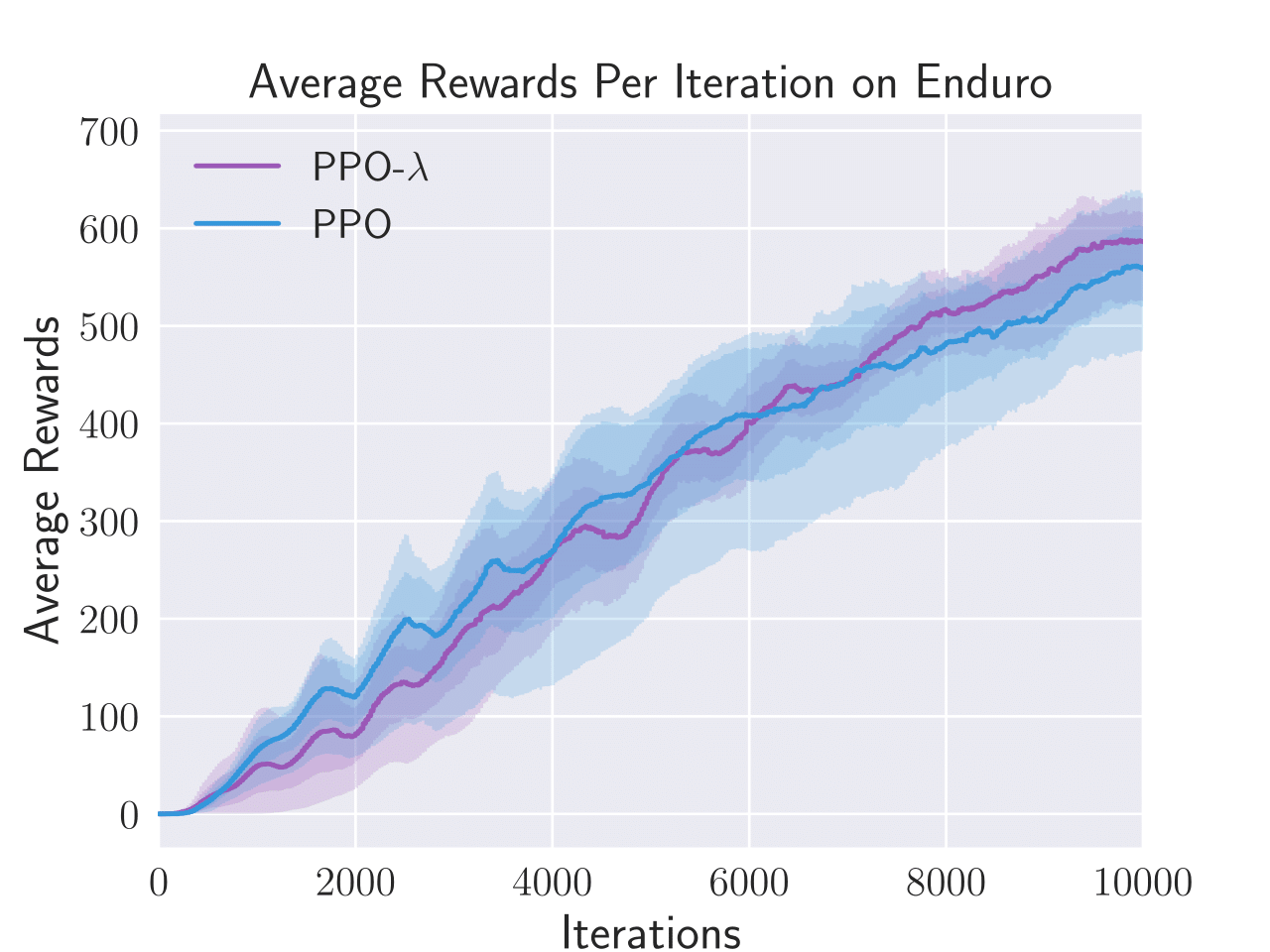}
        \subcaption{Enduro}
      \end{minipage}
      \begin{minipage}[t]{0.32\textwidth}
        \centering
        \includegraphics[width=\textwidth]{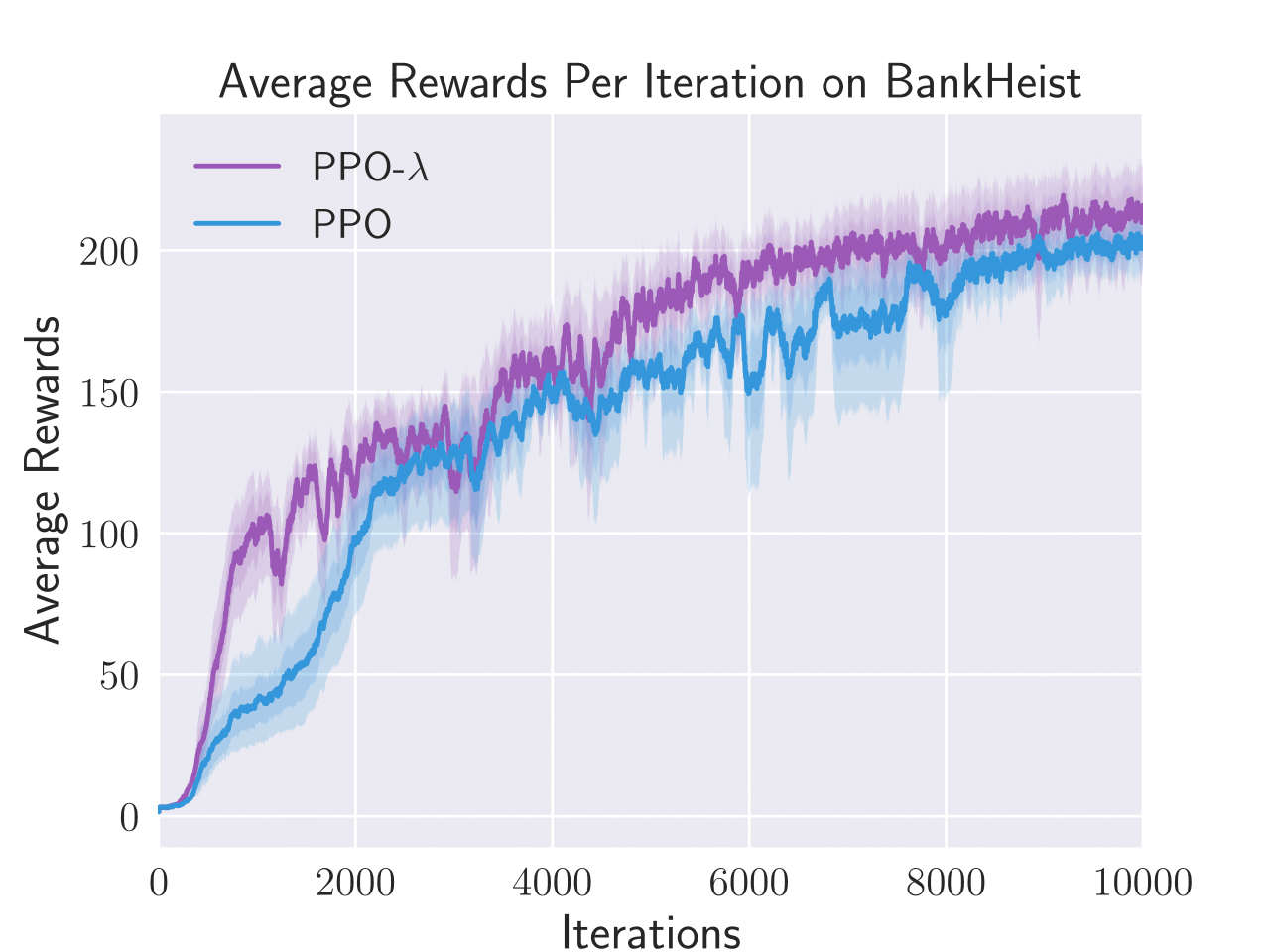}
        \subcaption{BankHeist}
      \end{minipage}%
      \begin{minipage}[t]{0.32\textwidth}
        \centering
        \includegraphics[width=\textwidth]{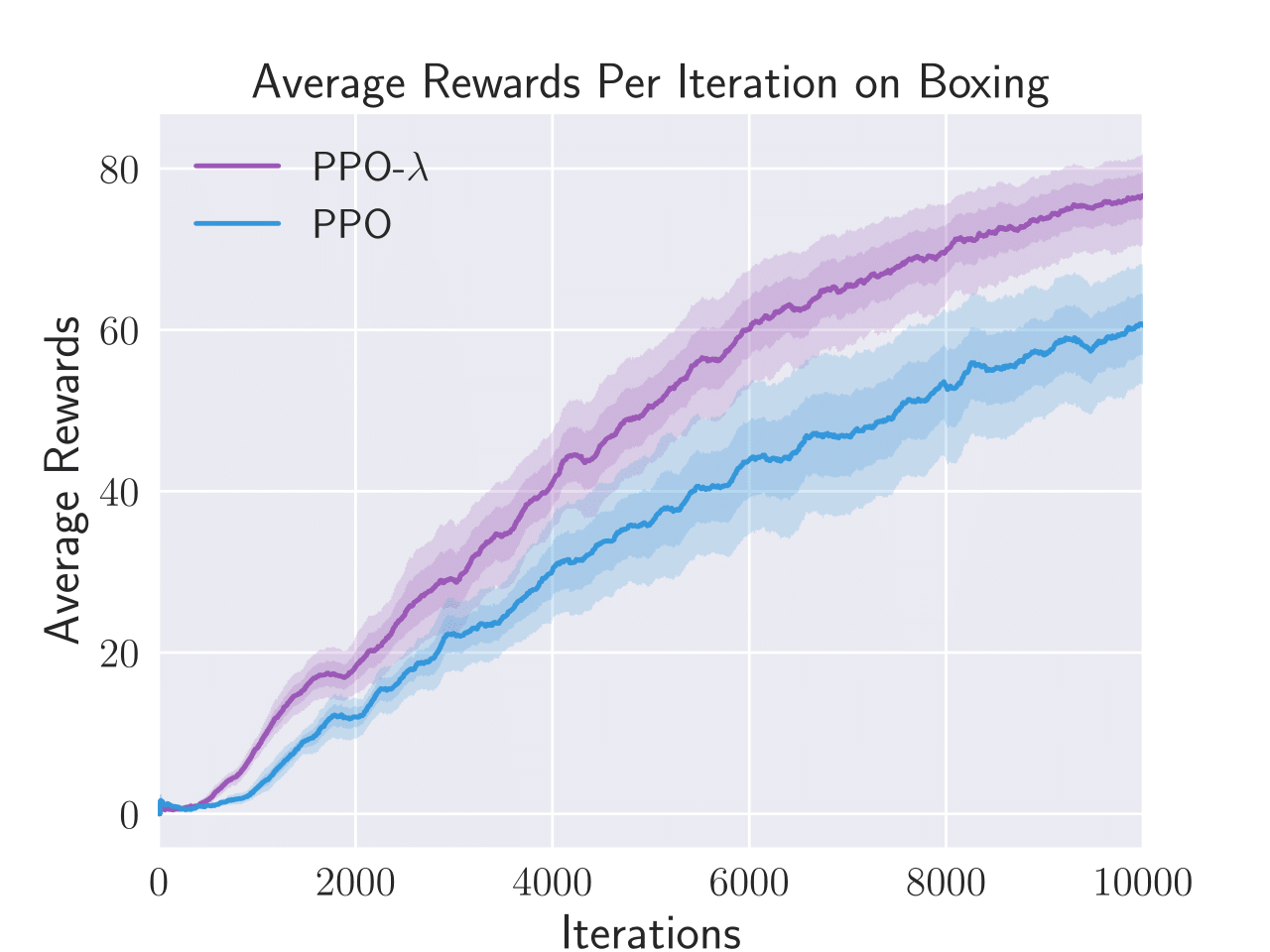}
        \subcaption{Boxing}
      \end{minipage}
      \\
      \begin{minipage}[t]{0.32\textwidth}
        \centering
        \includegraphics[width=\textwidth]{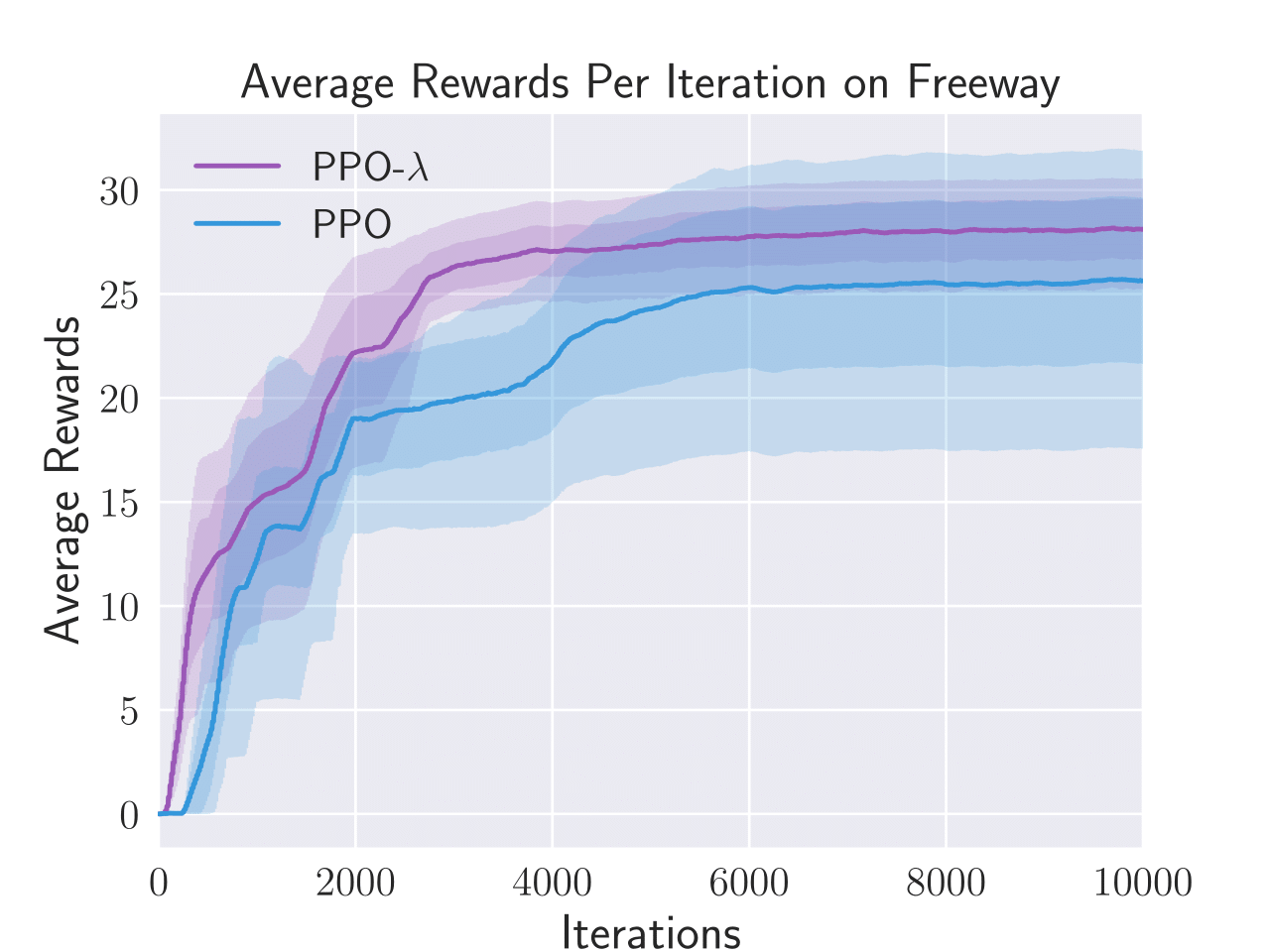}
        \subcaption{Freeway}
      \end{minipage}
      \begin{minipage}[t]{0.32\textwidth}
        \centering
        \includegraphics[width=\textwidth]{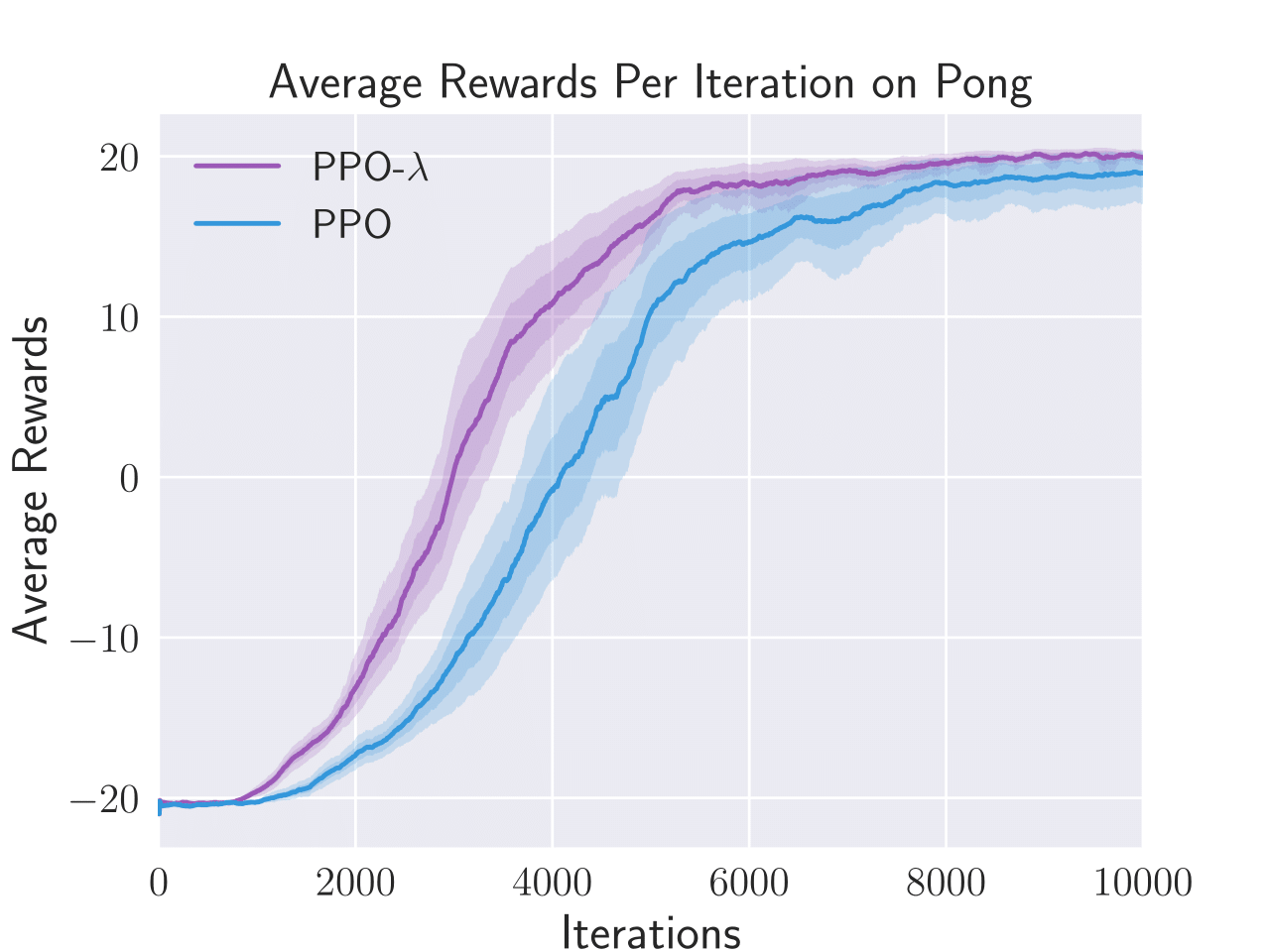}
        \subcaption{Pong}
      \end{minipage}%
      \begin{minipage}[t]{0.32\textwidth}
        \centering
        \includegraphics[width=\textwidth]{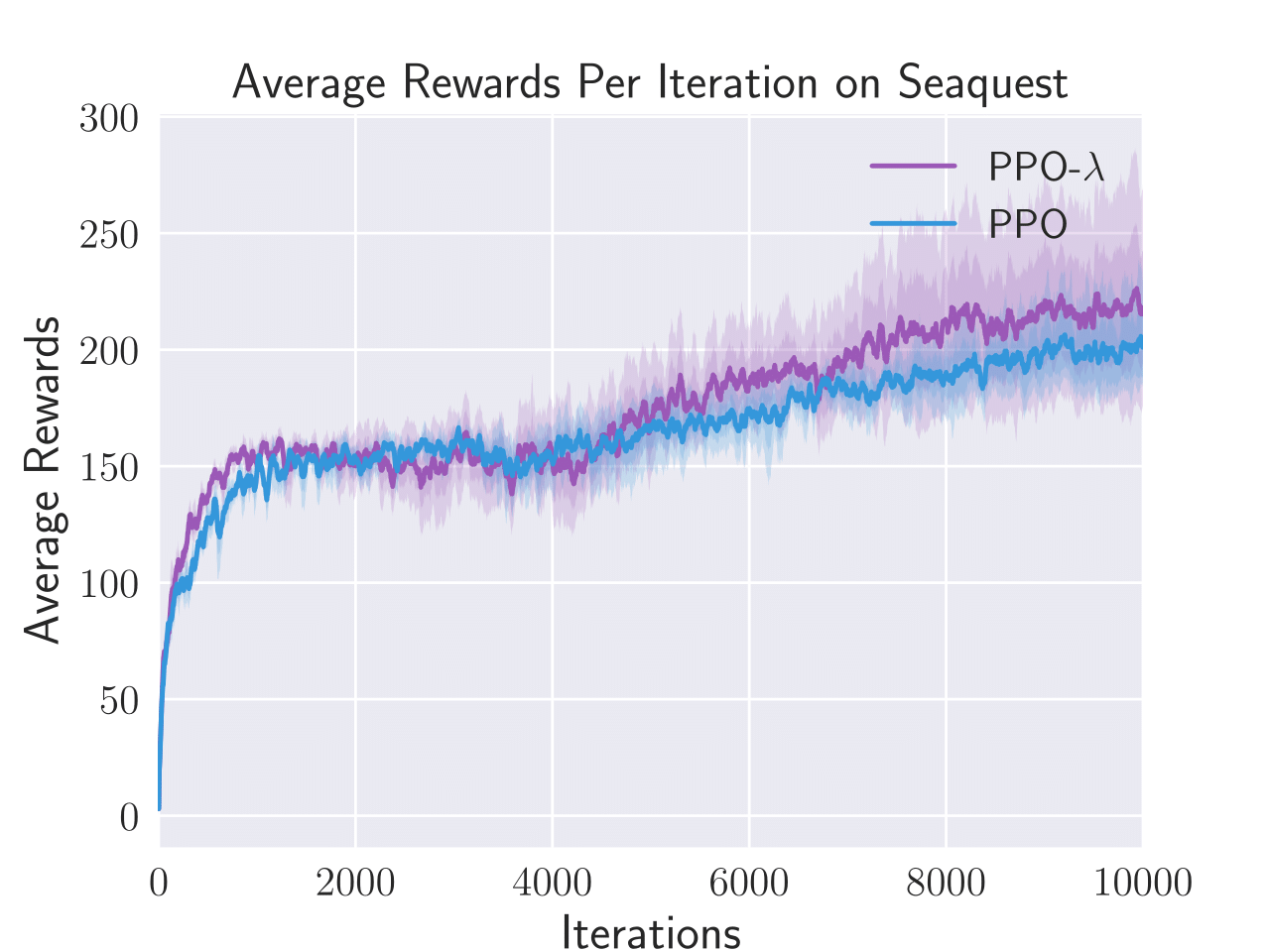}
        \subcaption{Seaquest}
      \end{minipage}
      \caption{Average total rewards per episode obtained by PPO-$\lambda$ and PPO on six Atari games, i.e. Enduro, BankHeist, Boxing, Freeway, Pong, and Seaquest.}
      \label{fig:learning_effectiveness}
  \end{figure*}

 \begin{figure*}[!t]
      \centering
      \begin{minipage}[t]{0.32\textwidth}
        \centering
        \includegraphics[width=\textwidth]{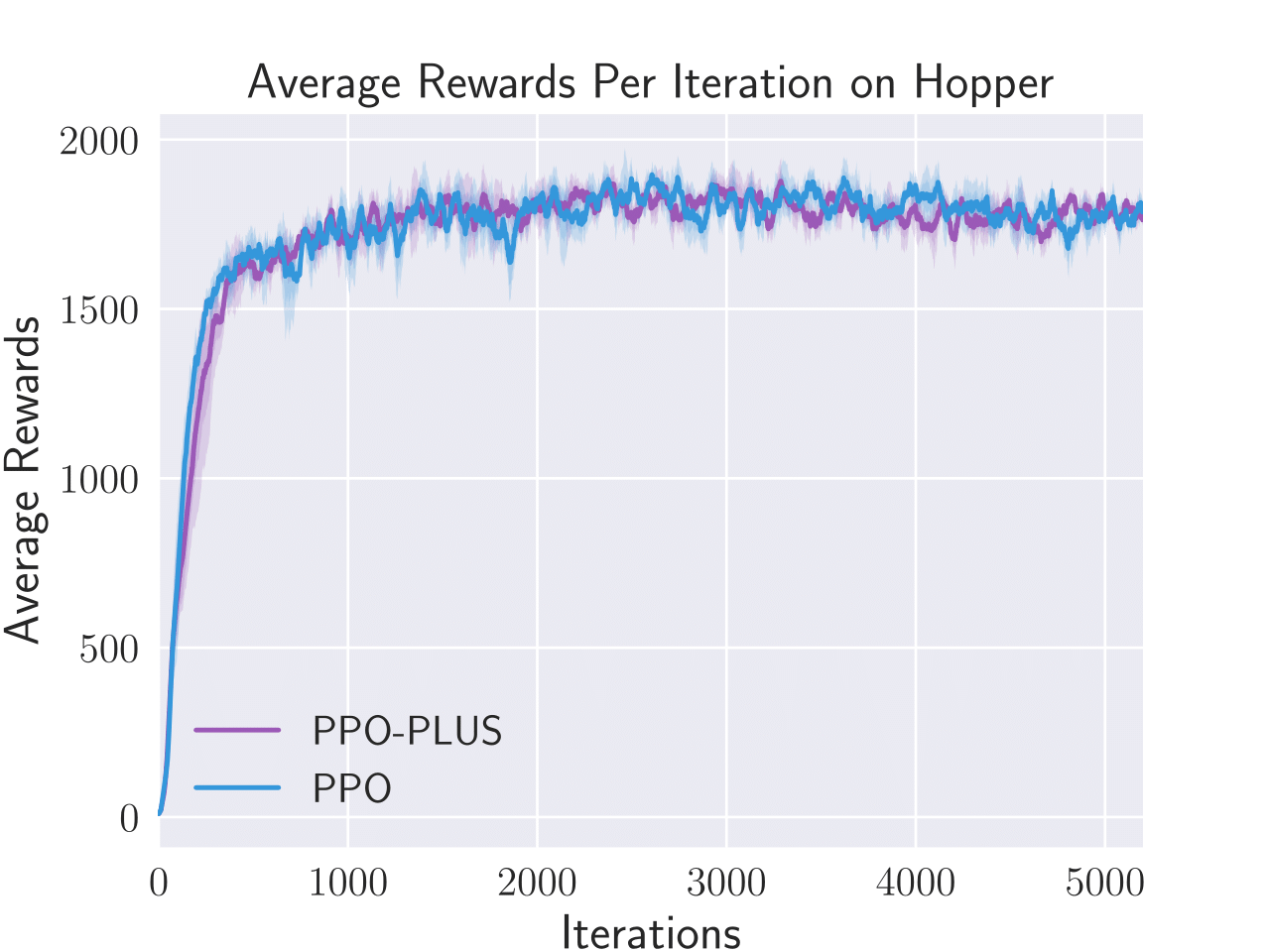}
        \subcaption{Hopper}
      \end{minipage}
      \begin{minipage}[t]{0.32\textwidth}
        \centering
        \includegraphics[width=\textwidth]{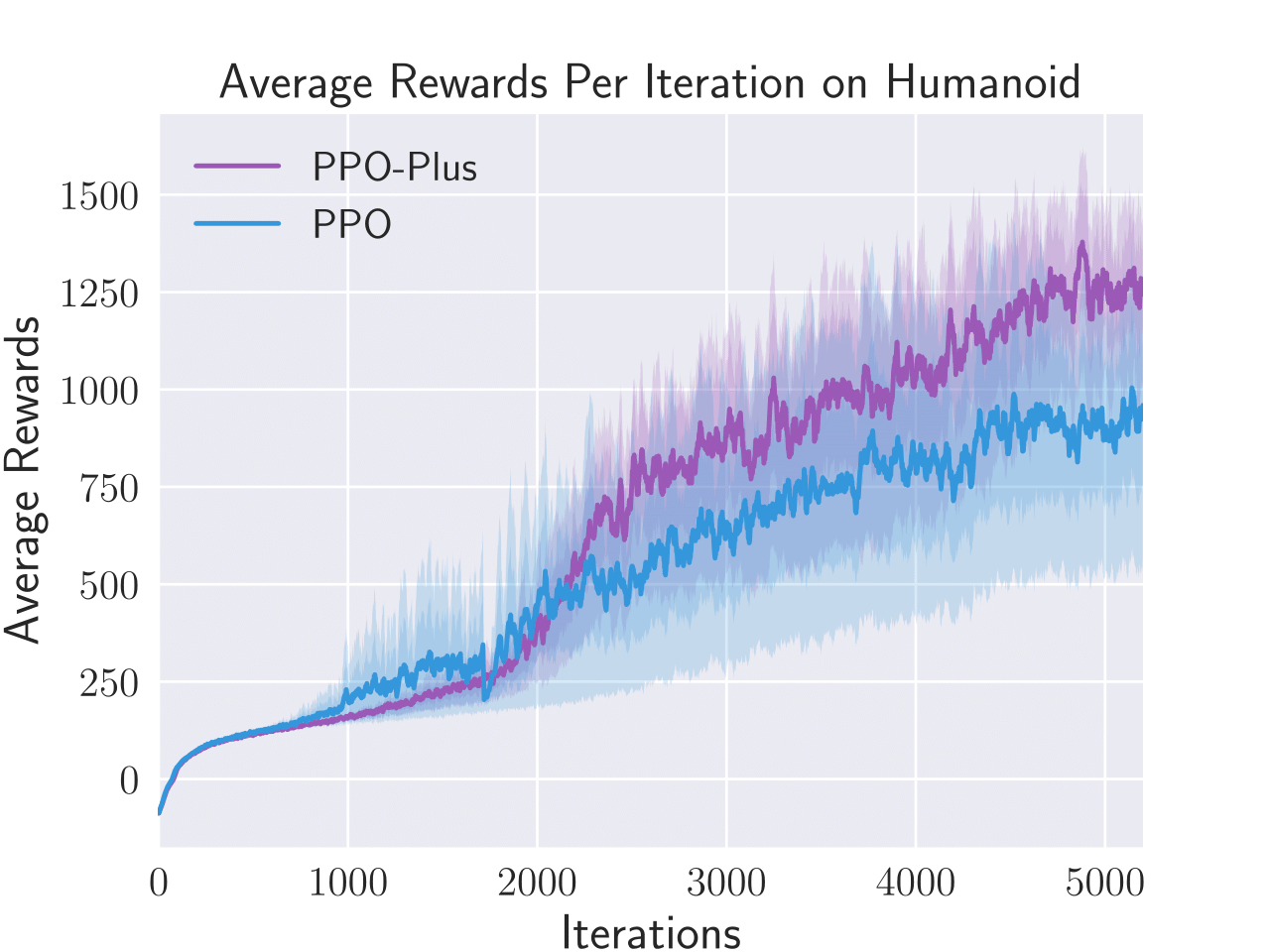}
        \subcaption{Humanoid}
      \end{minipage}
      \\
      \begin{minipage}[t]{0.32\textwidth}
        \centering
        \includegraphics[width=\textwidth]{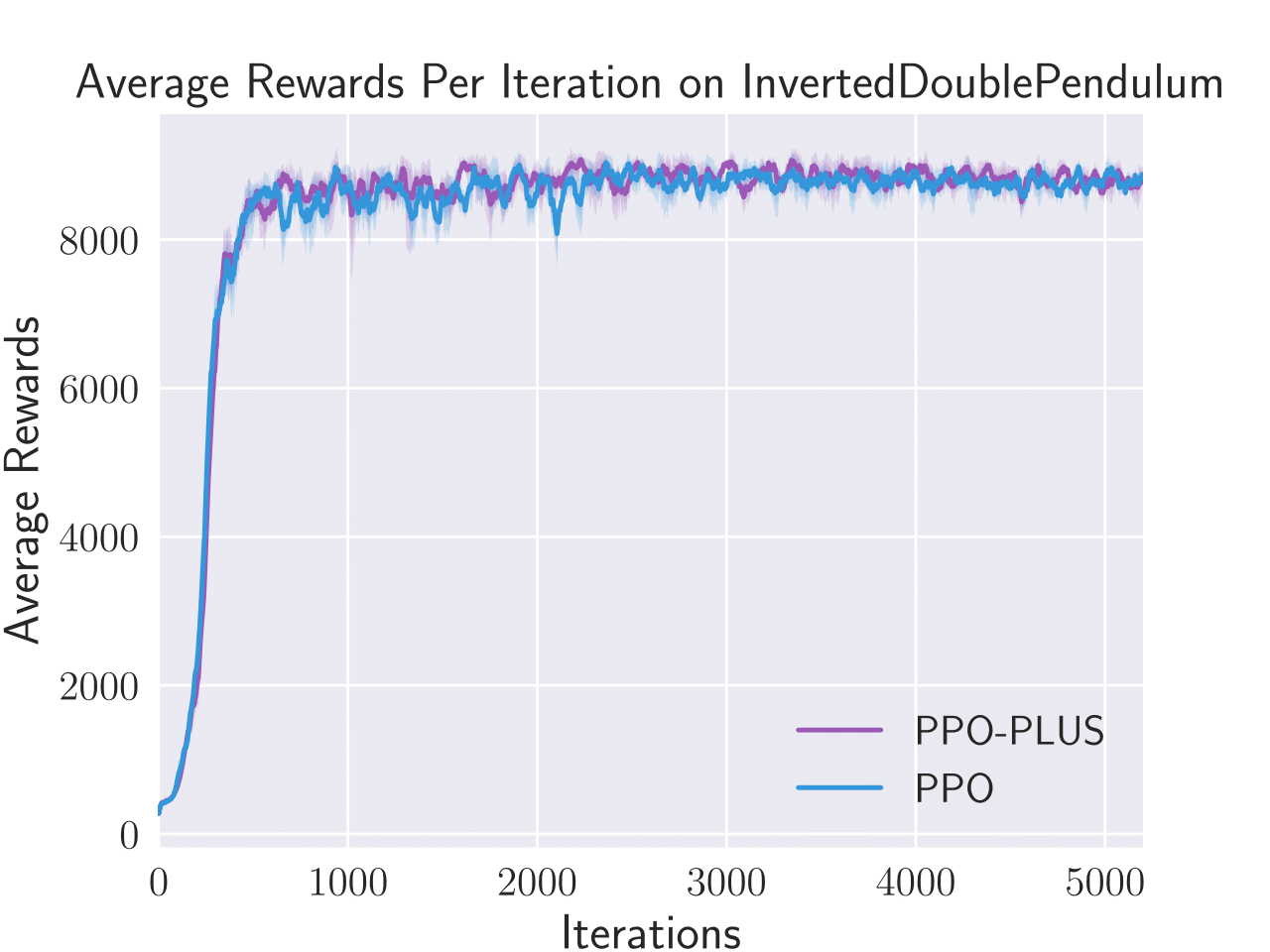}
        \subcaption{Inverted Double Pendulum}
      \end{minipage}
      \begin{minipage}[t]{0.32\textwidth}
        \centering
        \includegraphics[width=\textwidth]{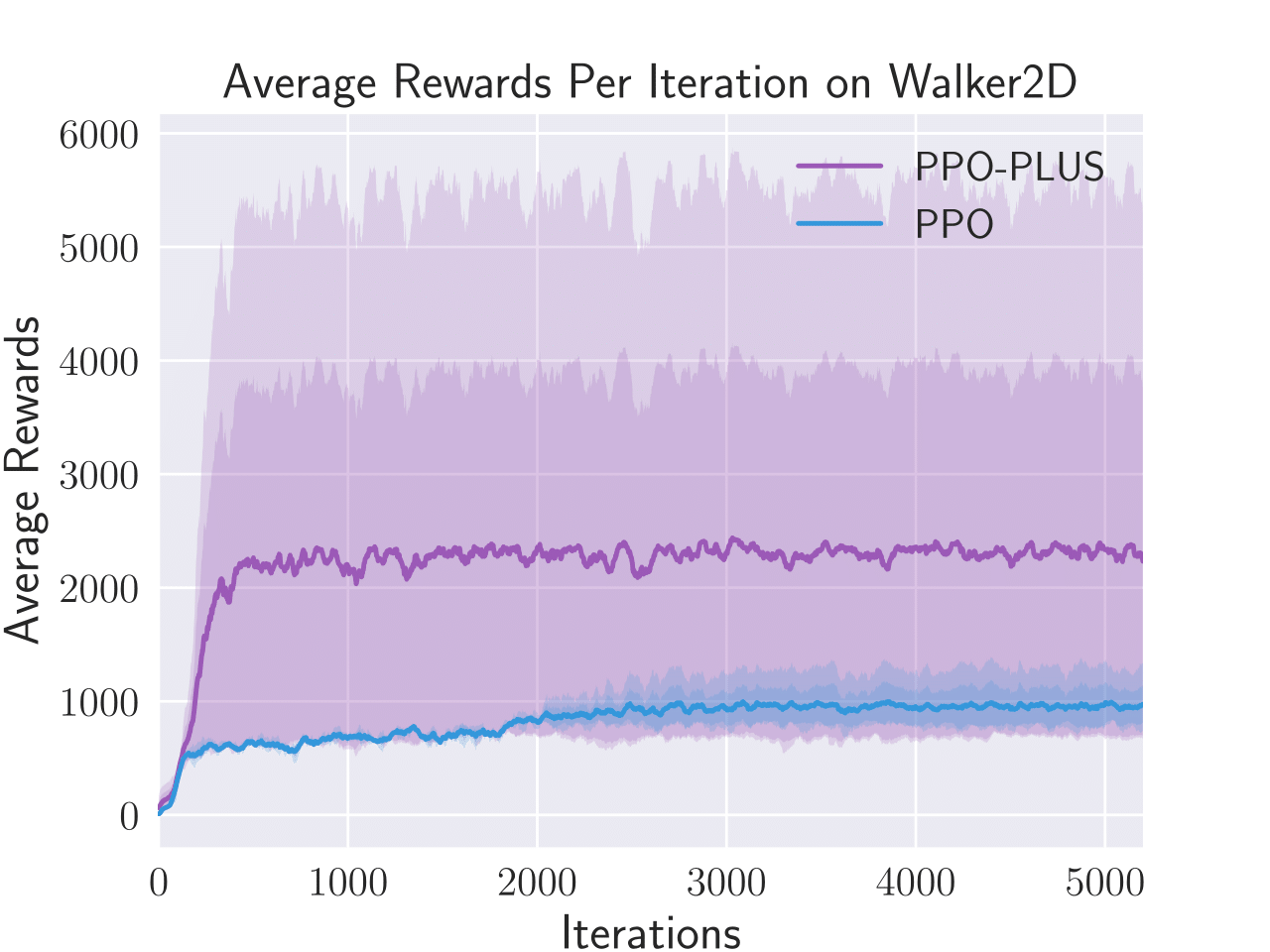}
        \subcaption{Walker2D}
      \end{minipage}
      \caption{Average total rewards per episode obtained by PPO-$\lambda$ and PPO on four benchmark control tasks, i.e. Hopper, Humanoid, Inverted Double Pendulum, and Walker2D.}
      \label{fig-con-learning-perf}
  \end{figure*}

\section{Empirical Analysis}
\label{sec-empirical_analysis}

In this section, we experimentally compare PPO-$\lambda$ and PPO on six benchmark Atari game playing tasks provided by OpenAI Gym \cite{Bellemare:2015vh,Brockman:2016wv} and four benchmark control tasks provided by the robotics RL environment of PyBullet \cite{pybullet}. First of all the performance of both algorithms will be examined based on the learning curves presented in Figure \ref{fig:learning_effectiveness} and Figure \ref{fig-con-learning-perf}. Afterwards, we will compare the sample efficiency of PPO-$\lambda$ and PPO by using the performance scores summarized in Table \ref{table-scoring-metric} and Table \ref{table-scoring-metric-control}.

In all our experiments, both algorithms adopt the same policy network architecture given in \cite{minh2015} for Atari game playing tasks and the same network architecture given in \cite{schulman2017,Duan2016} for benchmark control tasks. Meanwhile we follow strictly the hyperparameter settings used in \cite{schulman2017} for both PPO-$\lambda$ and PPO. The only exception is the minibatch size which is set to 256 in \cite{schulman2017} but equals to 128 in our experiments for Atari game playing tasks. This is because 8 actors will be employed by PPO-$\lambda$ and PPO while learning to play Atari games. Each actor will sample 128 time steps of data for every learning iteration. When the minibatch size is 128, one minibatch SGD update of the policy network will be performed by one actor, enabling easy parallelization.

In addition to the hyperparameters used in PPO, PPO-$\lambda$ requires an extra hyperparameter $\lambda$. In our experiments, $\lambda$ is set to $1$ initially without any parameter tuning. We also tested several different settings for $\lambda$, ranging from 0.1 to 5.0, but did not observe significant changes in performance. Based on the initial values, the actual value for $\lambda$ to be utilized in each learning iteration is further determined by (\ref{equ-lam-adj}). 

We obey primarily the default settings for all Atari games which are made available in the latest distribution of OpenAI baselines on Github. However, two minor changes have been introduced by us and will be briefly explained here. At each time step, following the DeepMind setting \cite{minh2015}, four consecutive frames of an Atari game must be combined together into one frame that serves subsequently as the input to the policy network. In OpenAI baselines, the combination is achieved by taking the maximum value of every pixel over the sampled frames. The snapshots of the first five sampled steps of the Pong game under this setting is presented in Figure \ref{fig:max_values}. Clearly, when the control paddle is moving fast, the visible size of the paddle in the combined frame will be larger than its actual size. As a result, an agent playing this game will not be able to know for sure that its observed paddle can catch an approaching ball. To address this issue, in our experiments, the value of each pixel in the combined frame is obtained by taking the average of the same pixel over sampled frames. The snapshots of the Pong game under this new setting can be found in Figure \ref{fig:mean_values}. Using Figure \ref{fig:mean_values}, the agent can easily determine which part of the visible paddle can catch a ball with certainly (i.e. the part of the paddle with the highest brightness).

We found that, in OpenAI baselines, every Atari game by default offers clipped rewards, i.e. the reward of performing any action is either -1, 0, or +1. Since  advantage function values will be normalized in both PPO-$\lambda$ and PPO, further clipping the rewards may not be necessary. Due to this reason, we do not clip rewards in all our experiments.

\begin{table}[]
\centering
\scalebox{0.6}{
\begin{tabular}{c|ccccccc}
\hline
Scoring Metric                     & Algorithms                         & BankHeist                & Boxing                  & Enduro                               & Freeway                 & Pong                    & Seaquest                  \\ \hline
\multirow{2}{*}{Fast Learning}     & PPO-$\lambda$                      & \textbf{160.29} & \textbf{47.88} & 332.93                     & \textbf{24.77} & \textbf{8.14}   & \textbf{177.99}  \\
                                   & \multicolumn{1}{c|}{PPO}           & 145.02          & 35.42          & \textbf{334.63}            & 21.34          & 3.82            & 168.74           \\ \hline
\multirow{2}{*}{Final Performance} & \multicolumn{1}{c|}{PPO-$\lambda$} & \textbf{211.54} & \textbf{77.34} & \textbf{643.82}            & \textbf{28.11} & \textbf{20.25} & \textbf{216.02} \\
                                   & \multicolumn{1}{c|}{PPO}           & 205.90          & 64.21          & \multicolumn{1}{l}{576.78} & 25.71          & 18.64           & 200.98
\end{tabular}}
\caption{Sample efficiency scores obtained by PPO-$\lambda$ and PPO on six Atari games.}
\label{table-scoring-metric}
\end{table}

\begin{table}[]
\centering
\scalebox{0.6}{
\begin{tabular}{c|c|cccc}
\hline
Scoring Metric                     & Algorithms    & Hopper                     & Humanoid                   & InvertedDoublePendulum     & Walker2D                   \\ \hline
\multirow{2}{*}{Fast Learning}     & PPO-$\lambda$ & \textbf{1718.75}          & \textbf{672.99}  & \textbf{8378.88} & \textbf{1954.56}  \\
                                   & PPO           & 1708.44 & 672.81           & 8319.28          & 830.23           \\ \hline
\multirow{2}{*}{Final Performance} & PPO-$\lambda$ & \textbf{1762.32}          & \textbf{1287.03} & \textbf{8869.84} & \textbf{2312.10} \\
                                   & PPO           & 1746.58 & 1232.64          & 8657.11          & 945.57
\end{tabular}}
\caption{Sample efficiency scores obtained by PPO-$\lambda$ and PPO on
four benchmark control tasks.}
\label{table-scoring-metric-control}
\end{table}

To compare the performance of PPO-$\lambda$ and PPO, we present the learning curves of the two algorithms on six Atari games in Figure \ref{fig:learning_effectiveness} and four benchmark control tasks in Figure \ref{fig-con-learning-perf}. As can be clearly seen in these figures, PPO-$\lambda$ can outperform PPO on five out of the six games (i.e. BankHeist, Boxing, Freeway, Pong and Seaquest) and two out of the four benchmark control tasks (i.e. Humanoid and Walker2D). On the Enduro game, PPO-$\lambda$ performed equally well as PPO. On some games such as Seaquest, the two algorithms exhibited similar performance within the first 5000 learning iterations. However PPO-$\lambda$ managed to achieve better performance towards the end of the learning process. On other games such as Boxing, Freeway and Pong, the performance differences can be witnessed shortly after learning started. It is also interesting to note that PPO-$\lambda$ did not perform clearly worse than PPO on any benchmark control tasks.

To compare PPO-$\lambda$ and PPO in terms of their sample efficiency, we adopt the two scoring metrics introduced in \cite{schulman2017}: (1) average reward per episode over the entire training period that measures fast learning and (2) average reward per episode over the last 10 training iterations that measures final performance. As evidenced in Table \ref{table-scoring-metric} and Table \ref{table-scoring-metric-control}, PPO-$\lambda$ is clearly more sample efficient than PPO on five out of six Atari games (with Enduro as the exception) and two out of four benchmark control tasks, based on their respective scores.

\section{Conclusions}
\label{sec-con}

Motivated by the usefulness of the simple clipping mechanism in PPO for efficient and effective RL, this paper aimed to develop new adaptive clipping method to further improve the performance of state-of-the-art deep RL algorithms. Particularly, through analyzing the clipped surrogate learning objective adopted by PPO, we found that repeated policy update in PPO may not adaptively improve learning performance in accordance with the importance of each sampled state. To address this issue, we introduced a new constrained policy learning problem at the level of individual states. Based on the solution to this problem, we proposed a new theoretical target for adaptive policy update, which enabled us to develop a new PPO-$\lambda$ algorithm. PPO-$\lambda$ is equally simple and efficient in design as PPO. By controlling a hyperparameter $\lambda$, we can effectively control policy update according to the significance of each sampled state and therefore enhance learning reliability. Moreover, our empirical study on six Atari games and four benchmark control tasks also showed that PPO-$\lambda$ can achieve better performance and higher sample efficiency than PPO in practice.

In the future,  it is interesting to study the effectiveness of PPO-$\lambda$ on many real-world applications including resource scheduling in large computer networks. It is also interesting to explore potential use of our adaptive clipping mechanism on A3C, ACKTR and other cutting-edge RL algorithms. Meanwhile more efforts must be spent to better understand the relationship between $\lambda$ and learning performance, which may give rise to more sample efficient algorithms.

\bibliographystyle{named}
\bibliography{citefile}

\begin{thebibliography}{}

\bibitem[\protect\citeauthoryear{Arulkumaran \bgroup \em et al.\egroup
  }{2017}]{arulkumaran2017}
K.~Arulkumaran, M.~P. Deisenroth, M.~Brundage, and A.~A. Bharath.
\newblock A brief survey of deep reinforcement learning.
\newblock {\em CoRR}, abs/1708.05866, 2017.

\bibitem[\protect\citeauthoryear{Barreto \bgroup \em et al.\egroup
  }{2017}]{barreto2017}
A.~Barreto, W.~Dabney, R.~Munos, J.~Hunt, T.~Schaul, D.~Silver, and H.~P. van
  Hasselt.
\newblock Successor features for transfer in reinforcement learning.
\newblock In {\em Advances in Neural Information Processing Systems}, pages
  4058--4068, 2017.

\bibitem[\protect\citeauthoryear{Bellemare \bgroup \em et al.\egroup
  }{2015}]{Bellemare:2015vh}
M.~G. Bellemare, Y.~Naddaf, J.~Veness, and M.~Bowling.
\newblock {The Arcade Learning Environment - An Evaluation Platform for General
  Agents (Extended Abstract).}
\newblock {\em IJCAI}, 2015.

\bibitem[\protect\citeauthoryear{Bhatnagar \bgroup \em et al.\egroup
  }{2009}]{bhatnagar2009}
S.~Bhatnagar, R.~S. Sutton, M.~Ghavamzadeh, and M.~Lee.
\newblock Natural actor-critic algorithms.
\newblock {\em Journal Automatica}, 45(11):2471--2482, 2009.

\bibitem[\protect\citeauthoryear{Brockman \bgroup \em et al.\egroup
  }{2016}]{Brockman:2016wv}
G.~Brockman, V.~Cheung, L.~Pettersson, J.~Schneider, J.~Schulman, J.~Tang, and
  W.~Zaremba.
\newblock {OpenAI Gym}.
\newblock {\em arXiv}, June 2016.

\bibitem[\protect\citeauthoryear{Coumans and Bai}{2018}]{pybullet}
E.~Coumans and Y.~Bai.
\newblock Pybullet, a python module for physics simulation for games, robotics
  and machine learning.
\newblock \url{http://pybullet.org}, 2018.

\bibitem[\protect\citeauthoryear{Duan \bgroup \em et al.\egroup
  }{2016}]{Duan2016}
Y.~Duan, X.~Chen, J.~Schulman, and P.~Abbeel.
\newblock {Benchmarking Deep Reinforcement Learning for Continuous Control}.
\newblock {\em arXiv}, 2016.

\bibitem[\protect\citeauthoryear{Gelfand \bgroup \em et al.\egroup
  }{2000}]{gelfand2000}
I.~M. Gelfand, R.~A. Silverman, et~al.
\newblock {\em Calculus of variations}.
\newblock Courier Corporation, 2000.

\bibitem[\protect\citeauthoryear{Gu \bgroup \em et al.\egroup }{2017}]{gu2017}
S.~Gu, T.~Lillicrap, Z.~Ghahramani, R.~E. Turner, B.~Sch{\"o}lkopf, and
  S.~Levine.
\newblock Interpolated policy gradient: Merging on-policy and off-policy
  gradient estimation for deep reinforcement learning.
\newblock {\em arXiv preprint arXiv:1706.00387}, 2017.

\bibitem[\protect\citeauthoryear{Haarnoja \bgroup \em et al.\egroup
  }{2017}]{haarnoja2017}
T.~Haarnoja, H.~Tang, P.~Abbeel, and S.~Levine.
\newblock Reinforcement learning with deep energy-based policies.
\newblock {\em arXiv preprint arXiv:1702.08165}, 2017.

\bibitem[\protect\citeauthoryear{Hausknecht and Stone}{2016}]{hausknecht2016}
M.~Hausknecht and P.~Stone.
\newblock On-policy vs. off-policy updates for deep reinforcement learning.
\newblock In {\em Deep Reinforcement Learning: Frontiers and Challenges, IJCAI
  2016 Workshop}, 2016.

\bibitem[\protect\citeauthoryear{Kingma and Ba}{2014}]{kingma2014}
D.~Kingma and J.~Ba.
\newblock Adam: A method for stochastic optimization.
\newblock {\em arXiv preprint arXiv:1412.6980}, 2014.

\bibitem[\protect\citeauthoryear{Lillicrap \bgroup \em et al.\egroup
  }{2015}]{lillicrap2015}
T.~Lillicrap, J.~J. Hunt, A.~Pritzel, N.~Heess, T.~Erez, Y.~Tassa, D.~Silver,
  and D.~Wierstra.
\newblock Continuous control with deep reinforcement learning.
\newblock {\em arXiv preprint arXiv:1509.02971}, 2015.

\bibitem[\protect\citeauthoryear{Mnih \bgroup \em et al.\egroup
  }{2015}]{minh2015}
V.~Mnih, K.~Kavukcuoglu, D.~Silver, A.~Rusu, J.~Veness, M.~G. Bellemare,
  A.~Graves, M.~Riedmiller, A.~K. Fidjeland, G.~Ostrovski, et~al.
\newblock Human-level control through deep reinforcement learning.
\newblock {\em Nature}, 518(7540):529--533, 2015.

\bibitem[\protect\citeauthoryear{Mnih \bgroup \em et al.\egroup
  }{2016}]{minh2016}
V.~Mnih, A.~P. Badia, M.~Mirza, A.~Graves, T.~Lillicrap, T.~Harley, D.~Silver,
  and K.~Kavukcuoglu.
\newblock Asynchronous methods for deep reinforcement learning.
\newblock In {\em International Conference on Machine Learning}, pages
  1928--1937, 2016.

\bibitem[\protect\citeauthoryear{Nachum \bgroup \em et al.\egroup
  }{2017}]{nachum2017}
O.~Nachum, M.~Norouzi, K.~Xu, and D.~Schuurmans.
\newblock Bridging the gap between value and policy based reinforcement
  learning.
\newblock In {\em Advances in Neural Information Processing Systems}, pages
  2772--2782, 2017.

\bibitem[\protect\citeauthoryear{Pollard}{2000}]{pollard2000}
D.~Pollard.
\newblock Asymptopia: an exposition of statistical asymptotic theory. 2000.
\newblock {\em URL http://www. stat. yale. edu/\~{} pollard/Books/Asymptopia},
  2000.

\bibitem[\protect\citeauthoryear{Schulman \bgroup \em et al.\egroup
  }{2015a}]{schulman2015}
J.~Schulman, S.~Levine, P.~Abbeel, M.~Jordan, and P.~Moritz.
\newblock Trust region policy optimization.
\newblock In {\em Proceedings of the 32nd International Conference on Machine
  Learning (ICML-15)}, pages 1889--1897, 2015.

\bibitem[\protect\citeauthoryear{Schulman \bgroup \em et al.\egroup
  }{2015b}]{schulman2015a}
J.~Schulman, P.~Moritz, S.~Levine, M.~Jordan, and P.~Abbeel.
\newblock High-dimensional continuous control using generalized advantage
  estimation.
\newblock {\em arXiv preprint arXiv:1506.02438}, 2015.

\bibitem[\protect\citeauthoryear{Schulman \bgroup \em et al.\egroup
  }{2017}]{schulman2017}
J.~Schulman, F.~Wolski, P.~Dhariwal, A.~Radford, and O.~Klimov.
\newblock Proximal policy optimization algorithms.
\newblock {\em arXiv preprint arXiv:1707.06347}, 2017.

\bibitem[\protect\citeauthoryear{Sutton and Barto}{1998}]{sutton1998}
R.~S. Sutton and A.~G. Barto.
\newblock {\em Reinforcement Learning: An Introduction}.
\newblock MIT Press, 1998.

\bibitem[\protect\citeauthoryear{Sutton \bgroup \em et al.\egroup
  }{2000}]{sutton2000}
R.~S. Sutton, D.~McAllester, S.~Singh, and Y.~Mansour.
\newblock Policy gradient methods for reinforcement learning with function
  approximation.
\newblock In {\em Advances in Neural Information Processing Systems 12 (NIPS
  1999)}, pages 1057--1063. MIT Press, 2000.

\bibitem[\protect\citeauthoryear{Szita and L{\"o}rincz}{2006}]{szita2006}
I.~Szita and A.~L{\"o}rincz.
\newblock Learning tetris using the noisy cross-entropy method.
\newblock {\em Learning}, 18(12), 2006.

\bibitem[\protect\citeauthoryear{Tang \bgroup \em et al.\egroup
  }{2017}]{tang2017}
H.~Tang, R.~Houthooft, D.~Foote, A.~Stooke, X.~Chen, Y.~Duan, J.~Schulman,
  F.~DeTurck, and P.~Abbeel.
\newblock \# exploration: A study of count-based exploration for deep
  reinforcement learning.
\newblock In {\em Advances in Neural Information Processing Systems}, pages
  2750--2759, 2017.

\bibitem[\protect\citeauthoryear{Vezhnevets \bgroup \em et al.\egroup
  }{2017}]{sasha2017feudal}
A.~S. Vezhnevets, S.~Osindero, T.~Schaul, N.~Heess, M.~Jaderberg, D.~Silver,
  and K.~Kavukcuoglu.
\newblock Feudal networks for hierarchical reinforcement learning.
\newblock {\em arXiv preprint arXiv:1703.01161}, 2017.

\bibitem[\protect\citeauthoryear{Wang \bgroup \em et al.\egroup
  }{2015}]{wang2015}
Z.~Wang, T.~Schaul, M.~Hessel, H.~Van Hasselt, M.~Lanctot, and N.~De Freitas.
\newblock Dueling network architectures for deep reinforcement learning.
\newblock {\em arXiv preprint arXiv:1511.06581}, 2015.

\bibitem[\protect\citeauthoryear{Wang \bgroup \em et al.\egroup
  }{2016}]{wang2016}
Z.~Wang, V.~Bapst, N.~Heess, V.~Mnih, R.~Munos, K.~Kavukcuoglu, and
  N.~de~Freitas.
\newblock Sample efficient actor-critic with experience replay.
\newblock {\em arXiv preprint arXiv:1611.01224}, 2016.

\bibitem[\protect\citeauthoryear{Wu \bgroup \em et al.\egroup }{2017}]{wu2017}
Y.~Wu, E.~Mansimov, R.~B. Grosse, S.~Liao, and J.~Ba.
\newblock Scalable trust-region method for deep reinforcement learning using
  kronecker-factored approximation.
\newblock In {\em Advances in Neural Information Processing Systems}, pages
  5285--5294, 2017.

\end{thebibliography}

\end{document}